\documentclass[aos,MSNbibl,seceqn,nameyear,dvips]{arximspdf}
\usepackage{mathbh}

\doi{10.1214/11-AOS961} 
\volume{40}
\issue{2}
\pubyear{2012}
\firstpage{639}
\lastpage{665}

\makeatletter

\newcommand{\cA}{\mathcal{A}}
\newcommand{\cC}{\mathcal{C}}
\newcommand{\cD}{\mathcal{D}}
\newcommand{\cE}{\mathcal{E}}
\newcommand{\cH}{\mathcal{H}}
\newcommand{\cK}{\mathcal{K}}
\newcommand{\cV}{\mathcal{V}}
\newcommand{\cX}{\mathcal{X}}
\newcommand{\cY}{\mathcal{Y}}
\newcommand{\cZ}{\mathcal{Z}}

\newcommand{\kE}{\mathfrak{E}}

\newcommand{\R}{{\mathbb{R}}}
\newcommand{\p}{{\mathbb{P}}}
\newcommand{\E}{{\mathbb{E}}}

\newcommand{\one}{{\mathbh{1}}}
\newcommand{\I}{{\mathbb{I}}}

\newcommand{\hh}{\mathsf{h}}

\newcommand{\argmin}{\mathop{\arg\min}}
\newcommand{\argmax}{\mathop{\arg\max}}
\newcommand{\ud}{\mathrm{d}}

\newtheorem{theorem}{Theorem}[section]
\newtheorem{cor}{Corollary}[section]
\newtheorem{lem}{Lemma}[section]

\newproclaim{cond}{Condition}

\makeatother

\begin{document}
\begin{frontmatter}

\title{Kullback--Leibler aggregation and misspecified generalized
linear models\thanksref{T1}}
\runtitle{Kullback--Leibler aggregation}

\thankstext{T1}{Supported in part by NSF Grants DMS-09-06424, DMS-10-53987
and AFOSR Grant A9550-08-1-0195.}

\begin{aug}
\author[A]{\fnms{Philippe} \snm{Rigollet}\corref{}\ead[label=e1]{rigollet@princeton.edu}}
\runauthor{P. Rigollet}
\affiliation{Princeton University}
\address[A]{Department of Operations Research \\
\quad and Financial Engineering\\
Princeton University\\
Princeton, New Jersey 08544\\
USA\\
\printead{e1}} 
\end{aug}

\received{\smonth{12} \syear{2010}}
\revised{\smonth{12} \syear{2011}}

%
\begin{abstract}
In a regression setup with deterministic design, we study the pure
aggregation problem and introduce a natural extension from the Gaussian
distribution to distributions in the exponential family. While this
extension bears strong connections with generalized linear models, it
does not require identifiability of the parameter or even that the
model on the systematic component is true. It is shown that this
problem can be solved by constrained and/or penalized likelihood
maximization and we derive sharp oracle inequalities that hold both in
expectation and with high probability. Finally all the bounds are
proved to be optimal in a minimax sense.
\end{abstract}

%
\begin{keyword}[class=AMS]
\kwd[Primary ]{62G08}
\kwd[; secondary ]{62J12}
\kwd{68T05}
\kwd{62F11}.
\end{keyword}
\begin{keyword}
\kwd{Aggregation}
\kwd{regression}
\kwd{classification}
\kwd{oracle inequalities}
\kwd{finite sample bounds}
\kwd{generalized linear models}
\kwd{logistic regression}
\kwd{minimax lower bounds}.
\end{keyword}

\end{frontmatter}

\section{Introduction}
\label{SECintro}

The last decade has witnessed a growing interest in the general problem
of \textit{aggregation}, which turned out to be a flexible way to
capture many statistical learning setups. Originally introduced in the
regression framework by \citet{Nem00} and \citet{JudNem00}
as an
extension of the problem of model selection, aggregation became a~mature statistical field with the papers of \citet{Tsy03} and
\citet{Yan04} where optimal rates of aggregation were derived.
Subsequent applications to density estimation [\citet{RigTsy07}] and
classification [\citet{BelSpo07}] constitute other illustrations
of the
generality and versatility of aggregation methods.

The general problem of
aggregation can be described as follows. Consider a finite family $\cH$
(hereafter called
\textit{dictionary}) of candidates for a certain
statistical task. Assume also that the dictionary $\cH$ belongs to a certain
linear space so that linear combinations of functions in $\cH$ remain
plausible candidates. Given a subset $\cC$ of the linear span
$\operatorname{span}(\cH)$ of $\cH$, the goal of aggregation is to mimic the best
element of $\cC$.

One salient feature of aggregation as opposed to standard statistical
modeling is that it does not rely on an underlying model. Indeed, the
goal is not to estimate the parameters of an underlying ``true'' model
but rather to construct an estimator that mimics the performance of the
best model in a given class, whether this model is true or not. From a
statistical analysis standpoint, this difference is significant since
performance cannot be measured in terms of parameters: there is no true
parameter. Rather, a stochastic optimization point of view is adopted.
If $R(\cdot)$ denotes a convex risk function, the goal pursued in
aggregation is to construct an \textit{aggregate estimator} $\hat h$
such that
%
%
\begin{equation}
\label{EQOIintro}
\E R(\hat h) \le C\min_{f \in\cC}R(f) +\varepsilon,
\end{equation}
where $\varepsilon$ is a small term that characterizes the performance of
the given aggregate~$\hat h$. As illustrated below, the remainder term
$\varepsilon$ is an explicit function of the size $M$ of the dictionary
and the sample size $n$ that shows the interplay between these two
fundamental parameters. Such oracle inequalities with optimal remainder
term $\varepsilon$ were originally derived by \citet{Yan00}
and \citet{Cat04} for model selection in the problems of density
estimation and Gaussian regression, respectively. They used a method,
called \textit{progressive mixture}, that was later extended to more
general stochastic optimization problems in \citet{JudRigTsy08}.
However, only bounds in expectation have been derived for this
estimator and it is argued in \citet{Aud08} that this estimator cannot
achieve optimal remainder terms with high probability. In the same
paper, Audibert suggests a different estimator that satisfies such an
oracle inequality with high probability at the cost of large constants
in the remainder term. One contribution (Theorem~\ref{THMSagP}) of
the present paper is to develop a new estimator that enjoys this
desirable property with small constants. We also study two other
aggregation problems: linear and convex aggregation.

When the model is misspecified, the minimum risk satisfies $\min_{f
\in
\cC}R(f)>0$, and it is therefore important to obtain a leading constant
$C=1$ in~(\ref{EQOIintro}). Many oracle inequalities with leading
constant term $C>1$ can be found in the literature for related
problems. \citet{Yan04} derives oracle inequalities with $C>1$ but
where the class $\cC=\cC_n$ actually depends on the sample size $n$ so
that $\min_{f \in\cC_n}R(f)$ goes to $0$ as $n$ goes to infinity under
additional regularity assumptions. In this paper, we focus on the
so-called \textit{pure} aggregation setup as defined by \citet
{Nem00} and
\citet{Tsy03} where the class~$\cC$ is fixed and remains very general.
As a result, we are only seeking oracle inequalities that have leading
constant $C=1$. Because they hold for finite $M$ and $n$, such oracle
inequalities are truly finite sample results.

The pure aggregation framework departs from the original problem of
aggregation, where the goal was to achieve adaptation by mimicking the
best of given estimators built from an independent sample. Thus a
typical aggregation procedure consists in splitting the sample in two
parts, using the first part to construct estimators and the second to
aggregate them [see, e.g., \citet{Lec07b}, \citet{RigTsy07}].
This procedure
relies heavily on the fact that the observations are identically
distributed, which is not the case in the fixed design regression
framework studied in the rest of the paper. It is worth mentioning that
in the case of model selection aggregation for Gaussian regression with
fixed design, the dictionary can be taken to be a family of projection
or even affine estimators built from the same sample. This specific
case has been investigated in more detail by
\citet{AlqLou11}, \citet{DalSal11}, \citet{RigTsy11},
but is beyond the scope of this paper.
Nevertheless, pure
aggregation, where the dictionary $\cH$ is deterministic, has grown
into a field of its own [see, e.g., \citet{BunTsyWeg07},
\citet{JudNem00}, \citet{JudRigTsy08}, \citet{Lou07},
\citet{Nem00}, \citet{Tsy03}]. In the case of
regression with fixed
design studied in this paper, the dictionary can be thought of as a
family of functions with minimal conditions that is expected to have
good approximation properties.

Pure aggregation turns out to be a stochastic optimization problem,
where the goal is to minimize an unknown risk function $R$ over a
certain set $\cC$. This paper is devoted to the case where the risk
function is given by the Kullback--Leibler divergence, and three
constraint sets that were introduced in \citet{Nem00} are
investigated.


We consider an extension of aggregation for Gaussian regression that
encompasses distributions for responses in a one-parameter exponential
family, with particular focus on the family of Bernoulli distributions
in order to cover binary \mbox{classification}. A natural measure of risk in
this problem is related to the Kullback--Leibler divergence between the
distribution of the actual observations and that of observations
generated from a given model. In a way, this extension is close to
generalized linear models [see, e.g., \citet{McCNel89}], which are
optimally solved by maximum likelihood estimation [see,
e.g., \citet{FahKau85}]. However, in the present aggregation
framework, it
is not assumed that there is one true model but we prove that maximum
likelihood estimators still perform almost as well as the optimal
solution of a suitable stochastic optimization problem. This
generalized framework encompasses logistic regression as a particular
case.

Throughout the paper, for any $x\in\R^n$, let $x_j$ denote its $j$th
coordinate. In other words, any vector $x \in\R^n$ can be written
$x=(x_1, \ldots, x_n)$. Similarly an $n \times M$ matrix $H$ has
coordinates $H_{i,j}, 1\le i\le n, 1\le j \le M$. The derivative of a
function $b\dvtx\R\to\R$ is
denoted by $b'$. For any real-valued function $f$, we denote by $\|f\|
_\infty={\sup_x} |f(x)| \in[0, \infty]$, its sup-norm. Finally, for any
two real numbers $x$ and $y$, we use the notation $x \wedge y=\min
(x,y)$ and $x \vee y=\max(x,y)$.

The paper is organized as follows. In the next section, we define the
problem of Kullback--Leibler aggregation, in the context of
misspecified generalized linear models. In particular, we exhibit a
natural measure of performance that suggests the use of constrained
likelihood maximization to solve it. Exact oracle inequalities, both in
expectation and with high probability, are gathered in Section~\ref{secmainUB} and their optimality for finite $M$ and $n$ is assessed in
Section~\ref{seclow}. These oracle inequalities for the case of large
$M$ are illustrated on a logistic regression problem, similar to the
problem of training a boosting algorithm, in Section~\ref{secexamples}.
Finally, Section~\ref{SECproofs} contains the proofs of the main
results together with useful properties on the concentration and the
moments of sums of random variables with distribution in an exponential family.

\section{Kullback--Leibler aggregation}
\label{secKLag}

\subsection{Setup and notation}

Let $x_1, \ldots, x_n$ be $n$ given points in a space $\cX$ and
consider the equivalence relation $\sim$ on the space of functions
$f\dvtx\cX\to\R$ that is defined such that $f\sim g$ if and only if
$f(x_i)=g(x_i)$ for all $i=1, \ldots, n$. Denote by $Q_{1:n}$ the
quotient space associated to this equivalence relation and define the
norm \mbox{$\|\cdot\|$} by
\[
\|f\|^2=\frac{1}{n}\sum_{i=1}^n f^2(x_i) ,\qquad f\in Q_{1:n} .
\]
Note that \mbox{$\|\cdot\|$} is a norm on the quotient space but only a
seminorm on the whole space of functions $f\dvtx\cX\to\R$.
%
%
%
%
In what follows, it will be useful to define the inner product
associated to \mbox{$\|\cdot\|$} by
\[
\langle f, g\rangle=\frac{1}{n}\sum_{i=1}^nf(x_i)g(x_i) .
\]
Using this inner product, we can also denote the average of a function
$f$ by $\langle f, \one\rangle$,
where $\one(\cdot)$ is the function in $Q_{1:n}$ that is identically
equal to 1.

Recall that a random variable $Y \in\R$ has distribution in a
(one-parameter) \textit{canonical exponential family} if it admits a
density with respect to a reference
measure on $\R$ given by
%
%
\begin{equation}
\label{EQexpdens2}
p(y;\theta)=\exp\biggl\{\frac{y\theta-b(\theta)}{a}+c(y) \biggr\}.
\end{equation}
%
A detailed treatment of exponential families of distributions together
with examples can be found in \citet{Bar78}, \citet{Bro86},
\citet{McCNel89} and in
\citet{LehCas98}. Several examples are also presented in
Section~\ref{secexamples} of the present paper.
It can be easily shown that if $Y$ admits a density given by~(\ref{EQexpdens2}), then
%
%
\begin{equation}
\label{EQvar}
\E[Y]=b'(\theta)  \quad\mbox{and}\quad
\operatorname{var}[Y]=ab''(\theta) .
\end{equation}
We assume hereafter that the distribution of $Y$ is not degenerate so
that~(\ref{EQvar}) ensures that $b$ is strictly convex and $b'$ is onto
its image space.

For any $g \in Q_{1:n}$, let $P_g$ denote the distribution of $n$
independent random variables $Y_1, \ldots, Y_n \in\cY\subset\R$ such
that $Y_i$ has density given by $p(y;\theta_i)$ where $\theta
_i=[b']^{-1}\circ g(x_i)$ so that $Y_i$ has expectation $g(x_i)$.

In this paper, we assume that we observe $n$ independent random
variables $Y_1, \ldots, Y_n \in\cY$ with joint distribution $\p=P_f$
for some unknown $f$. We denote by $\E$ the corresponding expectation.

\subsection{Aggregation and misspecified generalized linear models}
\label{sublinag}

When $\cX\subset\R^d$, generalized linear models (GLMs) assume that
the distribution of the observation $Y_i$ belongs to a given
exponential family with expectation\vspace*{1pt} $\E[Y_i]=f(x_i), i=1, \ldots, n$,
and that $l\circ f(x)=\beta^\top x$ where $l\dvtx\breve{\cY}\to\R$ is a
\textit{link function} and $\beta\in\R^d$ is the unknown parameter of
interest. A canonical choice for the link function is $l=[b']^{-1}$ and
in the rest of the paper, we study only this choice. In particular,
this canonical choice implies that $\theta_i=\beta^\top x_i$. While
GLMs allow more choices for the distribution of the response variable,
the modeling assumption $\theta_i=\beta^\top x_i$ is quite strong and
may be violated in practice. Aggregation offers a nice setup to study
the performance of estimators of $f$ even when this model is
misspecified.

Aggregation for the regression problem was introduced by \citet{Nem00}
and further developed by \citet{Tsy03} where the author considers a
regression problem with random design that has known distribution. We
now recall the main ideas of aggregation applied to the regression
problem, with emphasis on its difference with the linear regression
model. In the framework of the previous section, consider a finite
dictionary $\cH=\{h_1, \ldots, h_M \}$ such that $\|h_j\|$ is
finite and for any $\lambda\in\R^M$, let $\hh_\lambda$ denote the
linear combination of $h_j$'s defined by
%
%
\begin{equation}
\label{EQfflambda}
\hh_\lambda=\sum_{j=1}^M \lambda_j h_j .
\end{equation}
Assume that we observe $n$ independent random couples $(x_i,Y_i), i=1,
\ldots, n$, such that $\E[Y_i]=f(x_i)$.
The goal of \textit{aggregation} is to solve the following optimization problem:
%
%
\begin{equation}
\label{EQaggoptsto}
{\min_{\lambda\in\Lambda}}\|\hh_{\lambda}-f\|^2 ,
\end{equation}
where $\Lambda$ is a given subset of $\R^M$ and $f$ is unknown.
Previous papers on aggregation in the regression problem have focused
on three choices for the set $\Lambda$ corresponding to the three
different problems of aggregation originally introduced
by \citet{Nem00}. Optimal\vadjust{\goodbreak} rates of aggregation for these three problems
in the Gaussian regression setup can be found in \citet{Tsy03}.
\begin{longlist}
\item[\textsc{Model selection aggregation.}] The goal is to mimic the best
$h_j$ in the dictionary $\cH$. Therefore, we can choose $\Lambda$ to be
the finite set $\cV=\{e_1, \ldots, e_M\}$ formed by the $M$ vectors in
the canonical basis of $\R^M$. The optimal rate of model selection
aggregation in the Gaussian case is $(\log M)/n$.
\item[\textsc{Linear aggregation.}] The goal is to mimic the best linear
combination of the $h_j$'s in the dictionary $\cH$. Therefore, we can
choose $\Lambda$ to be whole space~$\R^M$. The optimal rate of linear
aggregation in the Gaussian case is~$M/n$.
\item[\textsc{Convex aggregation.}] The goal is to mimic the best convex
combination of the $h_j$'s in the dictionary $\cH$. Therefore, we can
choose $\Lambda$ to be the flat simplex of $\R^M$, denoted by
$\Lambda
_1^+$ and defined by
%
%
\begin{equation}
\label{EQell1ball}
\Lambda_1^+=\Biggl\{\lambda\in\R^M\dvtx\lambda_j\ge0, j=1, \ldots,
M,
\sum_{j=1}^M \lambda_j= 1\Biggr\}.
\end{equation}
%
The optimal rate of convex aggregation in the Gaussian case is
$(M/n)\wedge\sqrt{\log(1+M/\sqrt{n})/n}$.
\end{longlist}
%
In practice, the regression function $f$ is unknown and it is
impossible to perfectly solve~(\ref{EQaggoptsto}). Our goal is
therefore to recover an approximate solution of this problem in the
following sense. We wish
to construct an estimator~$\hat\lambda_n$ such that
%
%
\begin{equation}
\label{EQtermag}
\|\hh_{\hat\lambda_n} -f \|^2 - \min_{\lambda\in\Lambda}\|\hh
_{\lambda}-f\|^2
\end{equation}
is as small as possible. An inequality that provides an upper bound on
the (random) quantity in~(\ref{EQtermag}) in a certain probabilistic
sense is called \textit{oracle inequality}.

Observe that this is not a linear model since we do not assume that
the function $f$ is of the form $\hh_\lambda$ for some $\lambda\in
\R^M$. Rather, the bias term
${\min_{\lambda\in\Lambda}}\|\hh_{\lambda}-f\|^2$
may not vanish and the goal is to mimic the linear combination with the
smallest bias term.

The notion of Kullback--Leibler aggregation defined in the next
subsection broadens the scope of the above problem of aggregation to
encompass other distributions for $Y$.

\subsection{Kullback--Leibler aggregation}
\label{subGAP}

Recall that the ubiquitous squared norm \mbox{$\|\cdot\|^2$} as a measure of
performance for regression problems takes its roots in the Gaussian
regression model. The Kullback--Leibler divergence between two
probability distributions $P$ and $Q$ is defined by
\[
\cK(P\|Q)=\cases{
\displaystyle \int\log\biggl(\frac{\ud P}{\ud Q} \biggr)\,\ud P, &\quad if
$P \ll Q$,\vspace*{2pt}\cr
\infty, &\quad otherwise.}\vadjust{\goodbreak}
\]
Denote by $P_f$ the joint distribution of the observations $Y_i, i=1,
\ldots, n$. If $P_f$ denotes an $n$-variate Gaussian distribution with
mean $(f(x_1), \ldots, f(x_n))^\top$ and variance $\sigma^2I_n$, where
$I_n$ denotes the $n \times n$ identity matrix, then
$\cK(P_f\|\allowbreak
P_g)=\frac{n}{2\sigma^2}\|f-g\|^2$.
In order to allow an easier comparison between the results of this
paper and the literature, consider a normalized Kullback--Leibler
divergence defined by $\bar\cK(P_f\|P_g)=\cK(P_f\|P_g)/n$.
In the Gaussian regression setup, the quantity of interest in~(\ref{EQtermag}) can be written
%
%
\begin{equation}
\label{EQKLag}
\bar\cK(P_f\|P_{\hh_{\hat\lambda_n}})-\min_{\lambda\in\Lambda
}\bar
\cK(P_f\|P_{\hh_{\lambda}}) ,
\end{equation}
up to a multiplicative constant term equal to $2\sigma^2$.
Nevertheless, the quantity in~(\ref{EQKLag}) is meaningful for other
distributions in the exponential family.

Given a subset $\Lambda$ of $\R^M$, the goal of \textit
{Kullback--Leibler aggregation} (in short, KL-aggregation) is to
construct an estimator $\hat\lambda_n$ such that the \textit{excess-KL},
defined by
%
%
\begin{equation}
\label{EQKLag2}
\cE_{\mathrm{KL}}(\hh_{\hat\lambda_n}, \Lambda, \cH)=\bar\cK(P_f\|
P_{b'\circ\hh_{\hat\lambda_n}})-\inf_{\lambda\in\Lambda}\bar
\cK
(P_f\|P_{b' \circ\hh_{\lambda}}),
\end{equation}
is as small as possible.

Whereas KL-aggregation is a purely finite sample problem, it bears
connections with the asymptotic theory of model misspecification as
defined in \citet{Whi82}, following \citet{LeC53} and
\citet{Aka73}.
\citet{Whi82} proves that if the regression function $f$ is not
of the
form $f=b'\circ\hh_\lambda$ for some~$\lambda$ in the set of
parameters $\Lambda$, then under some identifiability and regularity
conditions, the maximum likelihood estimator converges to $\lambda^*$
defined by
\[
\lambda^*=\argmin_{\lambda\in\Lambda}\cK(P_f\|P_{b' \circ\hh
_{\lambda
}}) .
\]
Upper bounds on the excess-KL can be interpreted as finite sample
versions of those original results.

Note that assuming that $Y_i$ admits a density of the form~(\ref{EQexpdens2}) with known cumulant function $b(\cdot)$ is a strong
assumption unless $Y_i$ has Bernoulli distribution, in which case
identification of this distribution is trivial from the context of the
statistical experiment. We emphasize here that model misspecification
pertains only to the systematic component.

\section{Main results}
\label{secmainUB}

Let $\cZ=\{(x_1, Y_1), \ldots, (x_n, Y_n)\}$ be $n$ independent
observations and assume that for each $i$, the density of $Y_i$ is of
the form $p(y_i; \theta_i)$ as defined in~(\ref{EQexpdens2}) where
$\theta_i=[b']^{-1}\circ f(x_i)$. Then, we can write for any $\lambda
\in\R^M$,
%
%
\begin{equation}
\label{EQexpLL}
\cK(P_f\|P_{b' \circ\hh_{\lambda}})=-\frac{n}{a}(\langle f,
\hh
_\lambda\rangle-\langle b\circ\hh_\lambda, \one\rangle)
-\sum_{i=1}^n\E[c(Y_i)] +{\operatorname{Ent}}(P_f) ,\hspace*{-30pt}
\end{equation}
where ${\operatorname{Ent}}(P_f)$ denotes the entropy of $P_f$ and is defined by
\[
{\operatorname{Ent}}(P_f)=\sum_{i=1}^n\E\bigl[ \log\bigl(p\bigl(Y_i;
[b']^{-1}\circ f(x_i)\bigr) \bigr)\bigr] .
\]
Note that the term $-\sum_{i=1}^n\E[c(Y_i)]+{\operatorname{Ent}}(P_f)$ does not
depend on $\lambda$.

For estimators of the form $\hat\theta_i=\hh_\lambda(x_i)$, maximizing
the log-likelihood is equivalent to maximizing
%
%
\begin{equation}
\label{EQlogL}
\ell_n(\lambda)=\sum_{i=1}^n\{ Y_i \hh_\lambda(x_i)-\langle
b\circ
\hh_\lambda, \one\rangle\}
\end{equation}
over a certain set $\Lambda$ that depends on the problem at hand.

We now give bounds for the problem of KL-aggregation for the choices of~$\Lambda$ corresponding to the three problems of aggregation introduced
in the previous section. All proofs are gathered in Section~\ref{SECproofs} and rely on the following conditions, which can be easily
checked given the cumulant function~$b$.

%
\begin{cond}
\label{cond1}
The set of admissible parameters is $\Theta=\R$ and there exists a
positive constant $B^2$ such that
%
%
\begin{equation}
\label{EQUB2nddev}
\sup_{\theta\in\Theta} b''(\theta)\le B^2 .
\end{equation}
\end{cond}
%
%
\begin{cond}
\label{cond2}
We say that the couple $(\cH, \Lambda)$ satisfies Condition~\ref{cond2}
if there exists a positive constant $\kappa^2$ such that
\[
b''(\hh_{\lambda}(x))\ge\kappa^2 ,
\]
uniformly for all $x\in\cX$ and all $\lambda\in\Lambda$.
\end{cond}

Conditions~\ref{cond1} and~\ref{cond2} are discussed in the light of
several examples in Section~\ref{secexamples}. Condition~\ref{cond1} is
used only to ensure that the distributions of $Y_i$ have uniformly
bounded variances and sub-Gaussian tails, whereas Condition~\ref{cond2}
is a strong convexity condition that depends not only on the cumulant
function $b$ but also on the aggregation problem at hand that is
characterized by the couple $(\cH, \Lambda)$.

\subsection{Model selection aggregation}
\label{subMS} Recall that the goal of model selection aggregation is to
mimic a function $h_j$ such that $\cK(P_f\|P_{b' \circ h_j})\le
\cK(P_f\|P_{b' \circ h_k})$ for all $k\neq j$. A natural candidate
would be the function in the dictionary that maximizes the function
$\ell_n$ defined in~(\ref{EQlogL}) either over the finite set
$\cV=\{e_1, \ldots, e_M\}$ formed by the $M$ vectors in the canonical
basis of $\R^M$ or over its convex hull. However, it has been established [see, e.g.,
\citet{JudRigTsy08}, Lecu{\'e}
(\citeyear{Lec07b}), \citet{LecMen09}, \citet{RigTsy12}]
that such a choice is suboptimal in general. \citet{LecMen09}
proved that the maximum likelihood estimator on the flat simplex\vadjust{\goodbreak}
$\Lambda_1^+$ defined in Section~\ref{subcvx} is also suboptimal for
the problem of model selection. As a consequence, we resort to a
compromise between these two ideas and maximize a partially
interpolated log-likelihood. Define $\hat\lambda\in\Lambda_1^+$ to be
such that
%
%
\begin{equation}
\label{EQrhohat}
\hat\lambda\in\argmax_{\lambda\in\Lambda_1^+}\Biggl\{ \sum_{j=1}^M
\lambda_j\ell_n(e_j)+\ell_n(\lambda)\Biggr\}.
\end{equation}
Note that the criterion maximized in the above equation is the sum of
the log-likelihood and a linear interpolation of the values of the
log-likelihood at the vertices of the flat simplex. As argued above,
both of these terms are needed. Indeed, using only the linear
interpolation would lead us to choose~$\hat\lambda$ to be one of the
vertices of the simplex which, as mentioned above, is a~suboptimal choice.


%
%
\begin{theorem}
\label{THMSagE} Assume that Condition~\ref{cond1} holds and that
$(\cH,\Lambda_1^+)$ satisfies Condition~\ref{cond2}. Recall that
$\cV=\{e_1, \ldots, e_M\}$ is the finite set formed by the $M$ vectors
in the canonical basis of $\R^M$. Then, the aggregate $\hh_{\hat
\lambda}$ with $\hat\lambda$ defined in~(\ref{EQrhohat}) satisfies
%
%
\begin{equation}
\label{EQresMSagE1}
\E[\cE_{\mathrm{KL}}(\hh_{\hat\lambda}, \cV, \cH)]\le\frac{8
B^2}{\kappa^2}\frac{\log M}{n} .
\end{equation}
\end{theorem}

A similar result for $\hh_{\tilde\lambda}$ where $\tilde\lambda$ are
exponential weights was obtained by \citet{DalTsy07} for a different
class of regression problems with deterministic design under the
squared loss. For random design, \citet{JudRigTsy08} obtained
essentially the same results for the mirror averaging algorithm.
Also for random design, \citet{LecMen09} proposed a different estimator
to solve this problem and give for the first time a bound with high
probability with the optimal remainder term. Such a result was claimed
by \citet{Aud08} for a different estimator when the design is random.
Despite this recent effervescence, no bounds that hold with high
probability have been derived for the deterministic design case
considered here and the estimator proposed by \citet{LecMen09} is based
on a sample splitting argument that does not extend to deterministic
design. The next theorem aims at giving such an inequality for the
aggregate $\hh_{\hat\lambda}$.
%
%
\begin{theorem}
\label{THMSagP}
Assume that Condition~\ref{cond1} holds and that $(\cH,\Lambda_1^+)$
satisfies Condition~\ref{cond2}. Recall that $\cV=\{e_1, \ldots,
e_M\}$
is the finite set formed by the $M$ vectors in the canonical basis of
$\R^M$. Then, for any $\delta>0$, with probability $1-\delta$, the
aggregate $\hh_{\hat\lambda}$ with $\hat\lambda$ defined in~(\ref{EQrhohat}) satisfies
%
%
\begin{equation}
\cE_{\mathrm{KL}}(\hh_{\hat\lambda_n}, \cV, \cH)\le\frac{8
B^2}{\kappa
^2}\frac{\log(M/\delta)}{n} .
\end{equation}
\end{theorem}

The proofs of both theorems are gathered in Section~\ref{subprMS}.\vadjust{\goodbreak}

\subsection{Linear aggregation}
\label{sublin}
Let $\Lambda\subset\R^M$ be a closed convex set or $\R^M$ itself. The
\textit{maximum likelihood aggregate} over $\Lambda\subset\R^M$ is
uniquely defined as a~function in the quotient space $Q_{1:n}$ by the
linear combination $\hh_{\hat\lambda_n}$ with coefficients given by
%
%
\begin{equation}
\label{EQMLE}
\hat\lambda_n \in\argmax_{\lambda\in\Lambda}\ell_n(\lambda) .
\end{equation}
Note that both $\hat\lambda_n$ and $\lambda^*\in\argmin_{\lambda
\in
\Lambda}\cK(P_f\|P_{b' \circ\hh_{\lambda}})$
exist as soon as $\Lambda$ is a closed convex set [see \citet
{EkeTem99}, Chapter \textsc{ii}, Proposition~1.2]. Likewise, from the same
proposition, we find that if $\Lambda=\R^M$, Condition~\ref{cond2}
entails that both $\hat\lambda_n$ and $\lambda^*$ exist. Indeed, under
Condition~\ref{cond2}, the function $b$ is convex coercive and thus
both functionals
\[
\hh_\lambda\mapsto- \sum_{i=1}^n\{ Y_i \hh_\lambda
(x_i)-\langle
b\circ\hh_\lambda, \one\rangle\} \quad\mbox{and}\quad
\hh_\lambda\mapsto-\langle f, \hh_\lambda\rangle+\langle b\circ
\hh
_\lambda, \one\rangle
\]
are convex coercive. Thus,\vspace*{-1pt} the aggregates $\hh_{\lambda^*}$ and $\hh
_{\hat\lambda_n}$ are uniquely defined as functions in the quotient
space $Q_{1:n}$, even though $\lambda^*$ and $\hat\lambda_n$ may not
be unique.



We first extend the original results of \citet{Nem00} and
\citet{Tsy03} by providing bounds on the expected excess-KL, $\E
[\cE_{\mathrm{KL}}(\hh_{\hat\lambda_n}, \Lambda, \cH)]$ where
$\Lambda$ is either a closed convex set or $\Lambda=\R^M$, which
corresponds to the problem of linear aggregation.
%
%
\begin{theorem}
\label{THlinagE}
Let $\Lambda$ be a closed convex subset of $\R^M$ or $\R^M$ itself,
such that $(\cH, \Lambda)$ satisfies Condition~\ref{cond2}. If the
marginal variances satisfy $\E[Y_i-f(x_i)]^2\le\sigma^2$ for any $i=1,
\ldots, n$, then the maximum likelihood aggregate $\hh_{\hat\lambda
_n}$ over $\Lambda$ satisfies
%
%
\begin{eqnarray}
\label{EQresth1}
\E[\cE_{\mathrm{KL}}(\hh_{\hat\lambda_n}, \Lambda, \cH)]
&\le&\frac{2\sigma^2}{a\kappa^2}\frac{D}{n} ,\nonumber\\[-8pt]\\[-8pt]
\E\|\hh_{\hat\lambda_n}-\hh_{\lambda^*}\|^2&\le&
\frac{4\sigma^2}{\kappa^4}\frac{D}{n} ,
\nonumber
\end{eqnarray}
where $D\le M$ is the dimension of $\operatorname{span}(\cH)$ and $
\lambda^*\in\argmin_{\lambda\in\Lambda}\cK(P_f\|P_{b'\circ\hh
_{\lambda}})$.
\end{theorem}

Vectors $\lambda^*\in\argmin_{\lambda\in\Lambda}\cK(P_f\|
P_{b'\circ
\hh_{\lambda}})$ are oracles since they cannot be computed without the
knowledge of $P_f$. The oracle distribution $P_{b'\circ\hh_{\lambda
^*}}$ corresponds to the distribution of the form $P_{b'\circ\hh
_{\lambda}}, \lambda\in\Lambda$, that is the closest to the true
distribution $P_f$ in terms of Kullback--Leibler divergence.
Introducing this oracle allows us to assess the performance of the
maximum likelihood aggregate, without assuming that $P_f$ is of the
form $P_{b'\circ\hh_{\lambda}}$ for some $\lambda\in\Lambda$.
Note also that from~(\ref{EQvar}), the bounded variance condition $\E
[Y_i-f(x_i)]^2\le\sigma^2$ is a direct consequence of Condition~\ref{cond1} with $\sigma^2=aB^2$.

Theorem~\ref{THlinagE} is valid in expectation.
The following theorem shows that these bounds are not only valid in
expectation but also with high probability.\vadjust{\goodbreak}

%
\begin{theorem}
\label{THlinagP}
Let $\Lambda$ be a closed convex subset of $\R^M$ or $\R^M$ itself and
such that $(\cH, \Lambda)$ satisfies Condition~\ref{cond2}. Moreover,
let Condition~\ref{cond1} hold and let $D$ be the dimension of the
linear span of the dictionary $\cH=\{h_1 ,\ldots, h_M\}$. Then, for
any $\delta>0$, with probability $1-\delta$, the maximum likelihood
aggregate~$\hh_{\hat\lambda_n}$ over $\Lambda$ satisfies
%
%
\begin{eqnarray}
\cE_{\mathrm{KL}}(\hh_{\hat\lambda_n}, \Lambda, \cH)&\le&\frac
{8B^2}{\kappa
^2}\frac{D}{n}\log\biggl(\frac{4}{\delta}\biggr) ,\nonumber\\[-8pt]\\[-8pt]
\|\hh_{\hat\lambda_n}-\hh_{\lambda^*}\|^2&\le&\frac
{16aB^2}{\kappa
^4}\frac{D}{n}\log\biggl(\frac{4}{\delta}\biggr) ,
\nonumber
\end{eqnarray}
where $\lambda^*\in\argmin_{\lambda\in\Lambda}\cK(P_f\|
P_{b'\circ\hh
_{\lambda}})$.
\end{theorem}

We see that the price to pay to obtain bounds with high probability is
essentially the same as for the bounds in expectation up to an extra
multiplicative term of order $\log(1/\delta)$.
\subsection{Convex aggregation}
\label{subcvx}

In this subsection, we assume that $\Lambda\subset\Lambda_1^+$ is a~closed convex set. Note that both a maximum likelihood estimator $\hat
\lambda_n$ and an oracle $\lambda^*\in\argmin_{\lambda\in\Lambda
}\cK
(P_f\|P_{b' \circ\hh_{\lambda}})$
exist.

Recall that if $(\cH,\Lambda)$ satisfies Condition~\ref{cond2},
Theorems~\ref{THlinagE} and~\ref{THlinagP} also hold. The following
theorems ensure a better rate for the maximum likelihood aggregate
$\hh_{\hat\lambda_n}$ over $\Lambda$ when $D$, and thus $M$, becomes
much larger than $n$. It extends the problem of convex aggregation
defined by \citet{Nem00}, \citet{JudNem00} and \citet
{Tsy03} to the
case where the distribution of the response variables is not restricted
to be Gaussian.

%
\begin{theorem}
\label{THcvxagE}
Let $\Lambda$ be any closed convex subset of the flat simplex~$\Lambda
_1^+$ defined in~(\ref{EQell1ball}). Let Condition~\ref{cond1} hold and
assume that the dictionary~$\cH$ consists of functions satisfying $\|
h_j\|\le R$, for any $j=1, \ldots, M$ and some $R>0$.
Then, the maximum likelihood aggregate $\hh_{\hat\lambda_n}$ over
$\Lambda$ satisfies
%
%
\begin{equation}\label{EQresth22}
\E[\cE_{\mathrm{KL}}(\hh_{\hat\lambda_n}, \Lambda, \cH)]\le
RB\sqrt
{\frac{\log M}{an}} .
\end{equation}
Moreover, if $(\cH, \Lambda)$ satisfies Condition~\ref{cond2}, then
\[
\E\|\hh_{\hat\lambda_n}-\hh_{\lambda^*}\|^2\le\frac{2RB}{\kappa
^2}\sqrt{\frac{a\log M}{n}} ,
\]
where $\lambda^*\in\argmin_{\lambda\in\Lambda}\cK(P_f\|
P_{b'\circ\hh
_{\lambda}})$.
\end{theorem}

The bounds of Theorem~\ref{THcvxagE} also have a counterpart with high
probability as shown in the next theorem.
%
%
\begin{theorem}
\label{THcvxagP}
Let $\Lambda$ be any closed convex subset of the flat simplex~$\Lambda
_1^+$ defined in~(\ref{EQell1ball}). Fix\vadjust{\goodbreak} $M\ge3$, let Condition~\ref{cond1} hold and assume that the dictionary $\cH$ consists of functions
satisfying $\|h_j\|\le R$, for any $j=1, \ldots, M$ and some $R>0$.
Then, for any $\delta>0$, with probability $1-\delta$, the maximum
likelihood aggregate $\hh_{\hat\lambda_n}$ over $\Lambda$ satisfies
%
%
\begin{equation}\label{EQresth42}
\cE_{\mathrm{KL}}(\hh_{\hat\lambda_n}, \Lambda, \cH) \le RB\sqrt
{\frac
{2\log(M/\delta)}{an}}.
\end{equation}
Moreover, if $(\cH, \Lambda)$ satisfies Condition~\ref{cond2}, then on
the same event of probability $1-\delta$, it holds
%
%
\begin{equation}\label{EQresth44}
\|\hh_{\hat\lambda_n}-\hh_{\lambda^*}\|^2\le\frac{2RB}{\kappa
^2}\sqrt
{\frac{2a\log(M/\delta)}{n}} ,
\end{equation}
where $\lambda^*\in\argmin_{\lambda\in\Lambda}\cK(P_f\|
P_{b'\circ\hh
_{\lambda}})$.
\end{theorem}

This explicit logarithmic dependence in the dimension $M$ illustrates
the benefit of the $\ell_1$ constraint for high-dimensional problems.
\citet{RasWaiYu09} have obtained essentially the same result as
Theorem~\ref{THcvxagP} for the special case of Gaussian linear
regression. While their proof technique yields significantly larger
constants, they also cover the case of aggregation over $\ell_q$ balls
for $q<1$ explicitly. However, their result is limited to the linear
regression model where the regression function $f$ is of the form
$f=\hh_{\lambda^*}$ for some $\lambda^* \in\Lambda_1$, where
$\Lambda_1$ denotes the unit $\ell_1$ ball of $\R^M$.

Most of the existing bounds for convex aggregation hold for the
expected excess-KL.
Many papers provide bounds with high probability [see, e.g.,
Koltchinskii (\citeyear{Kol11}), \citet{Mas07}, \citet{MitGee09}
and references therein] but they typically do not hold for the
excess-KL itself but for a quantity related to
\[
\bar\cK(P_f\|P_{b'\circ\hh_{\hat\lambda_n}})-C\min_{\lambda\in
\Lambda}\bar\cK(P_f\|P_{b' \circ\hh_{\lambda}}) ,
\]
where $C>1$ is a constant. When the quantity $\min_{\lambda\in
\Lambda}\bar\cK(P_f\|P_{b' \circ\hh_{\lambda}})$ is not small enough,
such bounds can become uninformative. A notable exception is
Nemirovski et al. [(\citeyear{NemJudLan08}), Proposition 2.2]
where the authors derive a result
similar to Theorem~\ref{THcvxagP} under a different but similar set
of assumptions. Most importantly, their bounds do not hold for the
maximum likelihood estimator but for the output of a recursive
stochastic optimization algorithm.

\subsection{Discussion}

As mentioned before, it is worth noticing that the technique employed
in proving the bounds in expectation of the previous subsection yield
bounds with high probability at almost no extra cost.


We finally mention the question of \textit{persistence} posed by
Greenshtein and Ritov (\citeyear{GreRit04}) and further studied by \citet{Gre06}
and \citet{BarMenNee09}. In these papers, the goal is to find
performance bounds that explicitly depend on $n$, $M$ and the radius
$R$ of the~$\ell_1$ ball $R\Lambda_1$ when the functions of the
dictionary are scaled to have unit norm. Clearly, this is essentially
the same problem as ours if we choose the dictionary to be $\{0, Rh_1,
\ldots, Rh_M, -Rh_1, \ldots, -Rh_M\}$. More precisely, allowing~$M$ and
$R$ to depend on $n$, persistence asks the question of which regime
gives remainder terms that converge to $0$. While we do not pursue
directly this question, we can obtain such bounds for deterministic
design and show that the constrained maximum likelihood estimator on a
closed convex subset of the $\ell_1$ ball is persistent as long as
$R=R(n)=o(\sqrt{n/\log(M)})$. The original result
of \citet{GreRit04} in this sense allows only
$R=o([n/\log(M)]^{1/4})$ but when the design is random with
unknown distribution. The use of deterministic design in the present
paper makes the prediction task much easier. Indeed, a significant
amount of work to prove persistence has been made toward describing
general conditions on the distribution of the design to ensure
persistence at a rate $R=o(\sqrt{n/\log(M)})$, as
in \citet{Gre06} and \citet{BarMenNee09}.

\section{Optimal rates of aggregation}
\label{seclow}
In Section~\ref{secmainUB}, we have derived upper bounds for the
excess-risk both in expectation and with high probability under
appropriate conditions. The bounds in expectation can be summarized as
follows. For a given $\Lambda\subseteq\R^M$, there exists an
estimator $T_n$ such that its excess-KL satisfies
\[
\E[\bar\cK(P_f\|P_{T_n})]-\inf_{\lambda\in\Lambda
}\bar\cK
(P_f\|P_{b' \circ\hh_{\lambda}}) \le C\Delta_{n,M}( \Lambda) ,
\]
where $C>0$ and
%
%
\begin{equation}
\label{EQrecapUB}
\Delta_{n,M}( \Lambda)=\cases{
\displaystyle \frac{D}{n}\wedge\frac{\log M}{n} , \vspace*{2pt}\cr
\hphantom{\displaystyle \frac{D}{n},\qquad\hspace*{1.5pt}}\mbox{if
$\Lambda=\cV$}
\qquad\mbox{(model selection aggregation),}\vspace*{2pt}\cr
\displaystyle \frac{D}{n}, \mbox{\qquad if $\Lambda\subseteq\R^M$}
\qquad\mbox{(linear
aggregation),}\vspace*{2pt}\cr
\displaystyle \frac{D}{n}\wedge\sqrt{\frac{\log M}{n}}, \vspace*{2pt}\cr
\hphantom{\displaystyle \frac{D}{n},\qquad\hspace*{1.5pt}}\mbox{if
$\Lambda=\Lambda_1^+$} \qquad\mbox{(convex aggregation).}}\hspace*{-35pt}
\end{equation}
Here $D \le M \wedge n$ is the dimension of the linear span of the
dictionary $\cH$ and $\Lambda\subseteq\R^M$ means that $\Lambda$ is
either a closed convex subset of $\R^M$ or $\R^M$ itself.
Note that for model selection aggregation, the estimator that achieves
this rate is given by
$
T_n=b'\circ\hh_{\tilde\lambda_n}\I(D \ge\log M)+b'\circ\hh
_{\hat
\lambda_n}\I(D \le\log M) ,
$
where~$\tilde\lambda_n$ is defined in~(\ref{EQrhohat}), $\hh_{\hat
\lambda_n}$ is the maximum likelihood aggregate over $\Lambda_1^+$ and~$\I(\cdot)$ denotes the indicator function. Obviously, the lower bound
for linear aggregation does not hold for \textit{any} closed convex
subset of $\R^M$ since $\{0\}$ is such a~set and clearly $\Delta
_{n,M}(\{0\})\equiv0$. We will prove the lower bound on the unit~$\ell
_\infty$ ball defined by $
\Lambda_\infty=\{x\in\R^M\dvtx\max_{1\le j\le M}|x_j|\le
1\}$.

%
%
\begin{table}
\caption{Exponential families of distributions and constants in
Conditions \protect\ref{cond1} and \protect\ref{cond2} where $H_\infty$
is defined in~(\protect\ref{EQdefH}). [Source: McCullagh and Nelder
(\protect\citeyear{McCNel89})]}
\label{tableexpfam}
\begin{tabular*}{\tablewidth}{@{\extracolsep{\fill}}lccccccc@{}}
\hline
& $\bolds{\Theta}$ & $\bolds{\E(Y)}$ & $\bolds{a}$
& $\bolds{b(\theta)}$ & $\bolds{b''(\theta)}$ & $\bolds{B^2}$
& $\bolds{\kappa^2}$\\
\hline
Normal&$\R$& $\theta$ &$\sigma^2$&$\frac{\theta
^2}{2}$&$1$&$1$&$1$\\[5pt]
Bernoulli&$\R$&$\frac{e^\theta}{1+e^\theta}$&$1$&$\log(1+e^\theta)$&$
\frac{e^\theta}{(1+e^\theta)^2}$&$\frac{1}{4}$&$ \frac{e^{H_\infty
}}{(1+e^{H_\infty})^2}$\\[5pt]
Gamma &$(-\infty,0)$&$-\frac{1}{\theta}$ &$\frac{1}{\alpha
}$&$-\log
(-\theta)$& $1/\theta^2$& $\infty$ & $\frac{1}{H_\infty^2}$\\[5pt]
Negative binomial & $(0,\infty)$&$ \frac{r}{1-e^{\theta}}$& $1$&$
r\log
( \frac{e^{\theta}}{1-e^{\theta}})$& $ \frac{re^\theta
}{(1-e^\theta)^2}$&$\infty$&$ \frac{re^{H_\infty}}{
(1-e^{H_\infty
})^2}$\\[5pt]
Poisson&$\R$&$e^\theta$ &$1$ &$e^{\theta}$&$e^{\theta}$&$\infty
$&$e^{-H_\infty}$\\
\hline
\end{tabular*}
\end{table}

For linear and model selection aggregation, these rates are known to be
optimal in the Gaussian case where the design is random but with known\vadjust{\goodbreak}
distribution [\citet{Tsy03}] and where the design is
deterministic [\citet{RigTsy11}]. For convex aggregation, it has been
established by \citet{Tsy03} [see also \citet{RigTsy11}]
that the
optimal rate for Gaussian regression is of order
$\sqrt{\log(1+eM/\sqrt{n})/n}$, which is equivalent to the upper bounds
obtained in Theorems~\ref{THcvxagE}--\ref{THcvxagP} of the present
paper when $M \gg\sqrt n$ but is smaller in general. To obtain better
upper bounds, one may resort to more complicated, combinatorial
procedures such as the ones derived in the papers cited above but the
full description of this idea goes beyond the scope of this paper. Note
that in the case of bounded regression with quadratic risk and random
design, \citet{Lec11} recently proved that the constrained empirical
risk minimizer attains the optimal rate $\sqrt{\log(1+eM/\sqrt{n})/n}$
without any modification.

In this section, we prove that these rates are minimax optimal under
weaker conditions that are also satisfied by the Bernoulli
distribution. The notion of optimality for aggregation employed here is
a natural extension of the one introduced by \citet{Tsy03}.
Before stating the main result of this section, we need to introduce
the following definition. Fix $\kappa^2>0$ and let $\Gamma(\kappa^2)$
be the level set of the function $b''$ defined by
%
%
\begin{equation}
\label{EQdefGamma}
\Gamma(\kappa^2)=\{\theta\in\R\dvtx b''(\theta) \ge\kappa
^2\} .
\end{equation}
In the Gaussian case, it is clear from Table~\ref{tableexpfam} that
$\Gamma(\kappa^2)=\R$ for any $\kappa^2 \le1$. For the cumulant
function of the Bernoulli distribution, when $\kappa^2<1/4$, $\Gamma
(\kappa^2)$ is a compact symmetric interval given by
\[
\biggl[2\log\biggl(\frac{1-\sqrt{1-4\kappa^2}}{2\kappa}\biggr),
2\log
\biggl(\frac{1+\sqrt{1-4\kappa^2}}{2\kappa}\biggr)\biggr] .
\]
Furthermore, we have $\Gamma(1/4)=\{0\}$ and $\Gamma(\kappa
^2)=\varnothing$, for $\kappa^2>1/4$. In the next theorem, we assume
that for a given $\kappa^2>0$, $\Gamma(\kappa^2)$ is convex. This is
clearly the case when the cumulant function $b$ is such that $b''$ is
quasi-concave, that is, satisfies for any $\theta, \theta' \in\R, u
\in[0,1]$,
$b''(u\theta+ (1-u)\theta')\ge\min[b''(\theta), b''(\theta')]$.
This assumption is satisfied for the Gaussian and Bernoulli
distributions.\vadjust{\goodbreak}

Let $\bar\cD$ denote the class of dictionaries $\cH=\{h_1, \ldots,
h_M\}$ such that $\|h_j\|_\infty\le1$, $j=1,\ldots, M$. Moreover, for
any convex set $\Lambda\subseteq\R^M$, denote by $I(\Lambda)$ the
interval $[-H_\infty, H_\infty]$, where
%
%
\begin{equation}
\label{EQdefH}
H_\infty= H_\infty(\Lambda)={\sup_{\cH\in\bar\cD}\sup_{\lambda
\in
\Lambda}\sup_{x \in\cX} }|\hh_\lambda(x)| \in[0,\infty] .
\end{equation}
For example, we have
\[
I( \Lambda)=\cases{
[-1,1], &\quad if $\Lambda=\cV$\qquad\mbox{(model selection
aggregation),}\cr
\R, &\quad if $\Lambda=\R^M$\qquad\mbox{(linear aggregation),}\cr
[-1,1], &\quad if $\Lambda=\Lambda_1^+$\qquad\mbox{(convex
aggregation).}}
\]
To state the minimax lower bounds properly, we use the notation
\[
\mathfrak{E}_{\mathrm{KL}}(T_n, \Lambda, f, \cH)=\E[\bar\cK(P_f\|
P_{T_n})]-\inf_{\lambda\in\Lambda}\bar\cK(P_f\|P_{b' \circ
\hh
_{\lambda}}) ,
\]
that makes the dependence in the regression function $f$ explicit.
%
Finally, we denote by $E_{f}$ the expectation with respect to the
distribution $P_f$.
%
%
\begin{theorem}
\label{THlow}
Fix $M\ge2, n\ge1, D \ge1, \kappa^2>0$, and assume that
Condition~\ref{cond1} holds.
Moreover, assume that for a given set $\Lambda\subseteq\R^M$, we have
$I(\Lambda) \subset\Gamma(\kappa^2)$.
Then, there exists a dictionary $\cH\in\bar\cD$, with rank less than
$D$, and positive constants $c^*, \delta$ such that
%
%
\begin{equation}
\label{EQlowP}
\inf_{T_n}\sup_{\lambda\in\Lambda} P_{b' \circ\hh_\lambda}
\biggl[\kE
_{\mathrm{KL}}(T_n, \Lambda, b'\circ\hh_\lambda, \cH) > c_* \frac
{\kappa
^2}{2a}\Delta^*_{n,M}(\Lambda)\biggr]\ge\delta
\end{equation}
and
%
%
\begin{equation}
\label{EQlow}
\inf_{T_n}\sup_{\lambda\in\Lambda} E_{b' \circ\hh_\lambda}
[\kE
_{\mathrm{KL}}(T_n, \Lambda, b'\circ\hh_\lambda, \cH) ]\ge
\delta c_*
\frac{\kappa^2}{2a}\Delta^*_{n,M}(\Lambda) ,
\end{equation}
where the infimum is taken over all estimators and where
%
%
\begin{equation}
\label{EQoptUB}
\Delta^*_{n,M}( \Lambda)=\cases{
\displaystyle \frac{D}{n}\wedge\frac{\log M}{n}, &\quad if $\Lambda=\cV$,\vspace*{2pt}\cr
\displaystyle \frac{D}{n}, &\quad if $\Lambda\supset\Lambda_\infty(1)$,\vspace*{2pt}\cr
\displaystyle \frac{D}{n}\wedge\sqrt{\frac{\log(1+eM/\sqrt{n})}{n}}, &\quad
if $\Lambda=\Lambda_1^+$.}
\end{equation}
\end{theorem}

This theorem covers the Gaussian and the Bernoulli case for which
Condition~\ref{cond1} is satisfied. Lower bounds for aggregation in the
Gaussian case have already been proved in Rigollet and Tsybakov
[(\citeyear{RigTsy11}), Section 6]
in a~weaker sense. Indeed, we enforce here that $\cH\in\bar\cD$ and
has rank bounded by $D$, whereas \citet{RigTsy11} use unbounded
dictionaries with rank that may exceed $D$ by a logarithmic
multiplicative factor.

Observe that from~(\ref{EQlow}), the least favorable regression
functions are of the form $f=b'\circ\hh_\lambda, \lambda\in\Lambda$,
as it is the case for Gaussian aggregation [see, e.g.,
\citet{Tsy03}].\vadjust{\goodbreak}

A consequence of Theorem~\ref{THlow} is that the rates of convergence
obtained in Section~\ref{secmainUB}, both in expectation and with high
probability, cannot be improved without further assumptions except for
the logarithmic term of convex aggregation. The proof of Theorem~\ref{THlow} is provided in the supplementary material
[\citet{Rig12supp}].\vspace*{-2pt}

\section{Examples}\vspace*{-2pt}
\label{secexamples}

\subsection{Examples of exponential families}
This subsection is a reminder of the versatility of exponential
families of distributions and its goal is to illustrate Conditions~\ref{cond1} and~\ref{cond2} on some examples. Most of the material can be
found, for example, in \citet{McCNel89}. The form of the density
described in~(\ref{EQexpdens2}) is usually referred to as natural form.
We now recall that it already encompasses many different distributions.
Table~\ref{tableexpfam} gives examples of distributions that have such
a density. For distributions with several parameters, it is assumed
that all parameters but $\theta$ are known.
For the Normal and Gamma distributions, the reference measure is the
Lebesgue measure whereas for the Bernoulli, Negative binomial and
Poisson distributions, the reference measure is the counting measure on
$\mathbb{Z}$. For all these distributions, the cumulant function
$b(\cdot)$ is twice continuously differentiable.

Observe first that only the Normal and Bernoulli distributions satisfy
Condition~\ref{cond1}. Indeed, all other distributions in the table do
not have sub-Gaussian tails and therefore, we cannot use Lemma
\ref{lemchernoff} to control the deviations and moments of the sum of
independent random variables. Therefore, only Theorem~\ref{THlinagE}
applies to the remaining distributions even though direct computation
of the moments can yield results of the same type as Theorems~\ref{THcvxagE} and~\ref{THcvxagP} but with bounds that are larger by orders of
magnitude.

Another important message of Table~\ref{tableexpfam} is that the
constant $\kappa^2$ can depend on the constant $H_\infty$ defined
in~(\ref{EQdefH}).
Consequently the $L_2$ distance $\|\hh_{\hat\lambda_n}-\hh_{\lambda
^*}\|^2$ is affected by the constant $\kappa^2$ and thus by $H_\infty$.
However, the constant $B^2$ does not depend on $H_\infty$. Therefore,
the bounds on the excess-KL presented in Theorems~\ref{THcvxagE}
and~\ref{THcvxagP} hold without extra assumption of the dictionary. For
the Normal distribution, $\kappa^2=B^2=1$ regardless of the value~$H_\infty$, which makes it a particular case.\vspace*{-2pt}

\subsection{Bounds for logistic regression with a large dictionary}%

Let us now focus on the Bernoulli distribution. Recall that in the
setup of binary classification, we observe a collection of independent
random couples $(x_1, Y_1), \ldots,\allowbreak (x_n,Y_n)$ such that $Y_i \in\{
0,1\}$ has Bernoulli distribution with parameter~$f(x_i)$, $i=1, \ldots, n$.
As shown in the survey by \citet{BouBouLug05}, there exists a
tremendous amount of work in this topic and we will focus on the
so-called boosting type algorithms. A dictionary of base classifiers
$\cH=\{h_1, \ldots, h_M\}$, that is, functions taking values in
$[-1,1]$, is given and training a boosting algorithm consists in
combining them in such a way that $\hh_\lambda(x_i)$ predicts
$f(x_i)$ well.\vadjust{\goodbreak}

This part of the paper is mostly inspired by \citet{FriHasTib00}
who propose a statistical view of boosting following an original remark
of \citet{Bre99}. Specifically, they offer an interpretation of
the original \mbox{AdaBoost} algorithm introduced in \citet{FreSch96} as
a sequential optimization procedure that fits an extended additive
model for a particular choice of the loss function. Then they propose
to directly maximize the Bernoulli log-likelihood using quasi-Newton
optimization and derive a new algorithm called LogitBoost. Even though
we do not detail how maximization of the likelihood is performed,
LogitBoost aims at solving the same problem as the one studied here.
One difference here is that while extended additive models assume that
there exists $\lambda\in\Lambda\subset\R^M$ such that the
regression function is of the form $f=[b']^{-1}\circ\hh_\lambda$,
KL-aggregation does not. The paper of \citet{FriHasTib00} focuses
on the optimization side of the problem and does not contain finite
sample results. A~recent attempt to compensate for a lack of
statistical analysis can be found in \citet{MeaWyn08} and the many
discussions that it produced. We propose to contribute to this
discussion by illustrating some statistical aspects of LogitBoost based
on the rates derived in Section~\ref{secmainUB} and in particular, how
its performance depends on the size of the dictionary.

Given a convex subset $\Lambda\subset\R^M$ and a convex function
$\varphi\dvtx\R\to\R$, training a boosting algorithm, and more generally
a large margin classifier, consists in minimizing the risk function
defined by
\[
R_\varphi(\hh_\lambda)=\frac{1}{n}\sum_{i=1}^n\E[\varphi(-\tilde
Y_i\hh
_\lambda(x_i))]
\]
over $\lambda\in\Lambda$, where $\tilde Y_i=2Y_i -1 \in\{-1,1\}$.
It is not hard to show that minimizing the Kullback--Leibler
divergence $ \cK(P_f\|P_{b' \circ\hh_{\lambda}})$, is equivalent to choosing
%
%
\begin{equation}
\label{EQphilogit}
\varphi(x)=\frac{\log(1+e^{x})}{\log2} ,
\end{equation}
up to the normalizing constant $\log2$ that appears to ensure that
$\varphi(0)=1$.
For the choice of $\varphi$ defined in~(\ref{EQphilogit}), we have
\[
R_\varphi(\hh_\lambda)-\min_{\lambda\in\Lambda}R_\varphi(\hh
_\lambda
)=\frac{1}{\log2}\cE_{\mathrm{KL}}(\hh_\lambda, \Lambda, \cH) .
\]
In boosting algorithms, the size of the dictionary $M$ is much larger
than the sample size $n$ so that the results of Theorems~\ref{THlinagE} and~\ref{THlinagP} are useless and it is necessary to constrain
$\lambda$ to be in the rescaled flat simplex $R\Lambda_1^+$ so that
$H_\infty=R$. Given that for the Bernoulli distribution, we have $a=1,
B^2=1/4$, the constants in the main theorems can be explicitly computed
and in fact, they remain low. We can therefore apply Theorems~\ref{THcvxagE} and~\ref{THcvxagP} to obtain the following corollary that gives
oracle inequalities for the $\varphi$-risk $R_\varphi$, both in
expectation and with high probability. We focus on the case where $M$
is (much) larger than $ n$ as it is usually the case in boosting.
%
%
\begin{cor}
\label{corboosting}
Consider the boosting problem with a given dictionary of base
classifiers and let $\varphi$ be the convex function defined in~(\ref{EQphilogit}). Then, the maximum likelihood aggregate $\hh_{\hat
\lambda
_n}$ over the rescaled flat simplex $R\Lambda_1^+$, $R>0$, defined
in~(\ref{EQMLE}) satisfies
\[
\E[R_\varphi(\hh_{\hat\lambda_n})]\le\min_{\lambda\in
R\Lambda_1^+}R_\varphi(\hh_\lambda)+ \frac{R}{2\log2}\sqrt{\frac
{\log
M}{n}} .
\]
Moreover, for any $\delta>0$, with probability $1-\delta$, it holds
\[
R_\varphi(\hh_{\hat\lambda_n})\le\min_{\lambda\in R\Lambda
_1^+}R_\varphi(\hh_\lambda)+ \frac{R}{2\log2}\sqrt{\frac{2\log
(M/\delta)}{n}} .
\]
\end{cor}
%

\section{Proof of the main results}
\label{SECproofs}

In this section, we prove the main theorems. We begin by recalling some
properties of exponential families of distributions. While similar
results can be found in the literature, the results presented below are
tailored to our needs. In particular, the constants in the upper bounds
are explicit and kept as small as possible. In this section, for any
$\omega\in\ell_2(\R)$, denote by $|\omega|_2$ its $\ell_2$-norm.

\subsection{Some useful results on canonical exponential families}
\label{subexpfam}

Let $Y \in\R$ be a random variable with distribution in a canonical
exponential family that admits a density with respect to a reference
measure on $\R$ given by
%
%
\begin{equation}
\label{EQexpdens}
p(y;\theta)=\exp\biggl\{\frac{y\theta-b(\theta)}{a}+c(y) \biggr\}
,\qquad
\theta\in\R.
\end{equation}
It can be easily shown [see, e.g., \citet{LehCas98}, Theorem 5.10]
that the moment generating function of $Y$ is given by
%
%
\begin{equation}
\label{EQMGF}
\E[e^{tY} ]=e^{({b(\theta+at)-b(\theta)})/{a}} .
\end{equation}
Using~(\ref{EQMGF}) we can derive the Chernoff-type bounds presented in
the following lemma.
%
%
\begin{lem}
\label{lemchernoff}
$\!\!\!$Let $\omega=(\omega_1, \ldots, \omega_n) \in\R^n$ be a vector of
deterministic weights. Let $Y_1, \ldots, Y_n$ be independent random
variables such that $Y_i$ has density~$p(\cdot;\theta_i)$ defined
in~(\ref{EQexpdens}), $\theta_i \in\R$, $i=1, \ldots, n$, and define
the weighted sum $S_n^\omega=\sum_{i=1}^n \omega_i Y_i$. Assume that
Condition~\ref{cond1} holds. Then the following inequalities hold:
%
%
\begin{eqnarray}
\label{EQmgf}
\E\bigl[ \exp\bigl(s|S_n^\omega-\E(S_n^\omega)|\bigr)\bigr] &\le&\exp
\biggl(\frac
{s^2B^2a|\omega|_2^2}{2}\biggr) ,
\\
%
%
\label{EQdeviation}
\p[|S_n^\omega-\E(S_n^\omega)| > t ]&\le&2\exp
\biggl(-\frac
{t^2}{2aB^2|\omega|_2^2}\biggr) ,\vadjust{\goodbreak}
\end{eqnarray}
and for any $r\ge0$, we have
%
%
\begin{equation}
\label{EQmoment}
\E|S_n^\omega-\E(S_n^\omega)|^r \le C_r|\omega|_2^{r} ,
\end{equation}
where $C_r=r(2aB^2)^{r/2}\Gamma(r/2)$ and $\Gamma(\cdot)$ denotes the
Gamma function.
\end{lem}
%
%
\begin{pf}
Using, respectively,~(\ref{EQMGF}),~(\ref{EQvar}) and~(\ref{EQUB2nddev}),
we get
\begin{eqnarray*}
\E\bigl[ \exp\bigl(s\bigl(S_n^\omega-\E(S_n^\omega)\bigr)\bigr)\bigr]&=&\exp
\Biggl(\frac
{1}{a}\sum_{i=1}^n [b(\theta_i+as\omega_i)-b(\theta_i)-as\omega
_ib'(\theta_i)] \Biggr)\\
&\le&\exp\biggl(\frac{s^2B^2a|\omega|_2^2}{2}\biggr) .
\end{eqnarray*}
The same inequality holds with $s$ replaced by $-s$ so~(\ref{EQmgf}) holds.

The proof of~(\ref{EQdeviation}) follows from~(\ref{EQmgf}) together
with a Chernoff bound.
Next, note that
\[
\E|S_n^\omega-\E(S_n^\omega)|^r =\int_0^\infty\p\bigl(|S_n^\omega-\E
(S_n^\omega)|>t^{1/r}\bigr)\,\ud t\le2\int_0^\infty\exp
\biggl(-\frac{t^{2/r}}{2aB^2|\omega|_2^2}\biggr)\,\ud t ,
\]
where we used~(\ref{EQdeviation}) in the last inequality. Using a
change of variable, it is not hard to see that this bound yields~(\ref{EQmoment}).
\end{pf}
%


\subsection{\texorpdfstring{Proof of Theorems \protect\ref{THMSagE} and \protect\ref{THMSagP}}{Proof of Theorems 3.1 and 3.2}}
\label{subprMS}

According to~(\ref{EQexpLL}), minimizing $\lambda\mapsto\cK(P_f\|P_{b'
\circ\hh_{\lambda}})$ is equivalent to maximizing $\lambda\mapsto
L(\lambda)$ where
%
%
\begin{equation}
\label{EQdefL}
L(\lambda)=\langle f, \hh_\lambda\rangle-\langle b\circ\hh
_\lambda,
\one\rangle.
\end{equation}
Note that for any $\Lambda\subset\R^M$, the set of optimal solutions
$\Lambda^*$ satisfies
\[
\Lambda^*=\argmin_{\lambda\in\Lambda}\cK(P_f\|P_{b'\circ\hh
_{\lambda
}})=\argmax_{\lambda\in\Lambda}L(\lambda) .
\]
Moreover, for any $\lambda\in\Lambda, \lambda^* \in\Lambda^*$, we have
%
%
\begin{equation}
\label{EQLKL}
L({\lambda^*})-L(\lambda)=a\cE_{\mathrm{KL}}(\hh_{ \lambda}, \Lambda,
\cH) .
\end{equation}
For any fixed $\lambda\in\Lambda_1^+$, define the following quantities:
\begin{eqnarray*}
S_n(\lambda)&=&\sum_{j=1}^M \lambda_j\ell_n(e_j)+\ell_n(\lambda)
,\\
S(\lambda)& = &n\sum_{j=1}^M\lambda_j L(e_j)+nL(\lambda)\vadjust{\goodbreak}
\end{eqnarray*}
and observe that $S(\lambda)=\E[S_n(\lambda)]$ and that for any
$\lambda\in\Lambda_1^+$,
\[
S_n(\lambda)-S(\lambda)=2\sum_{i=1}^n \bigl(Y_i - f(x_i)\bigr)\hh
_\lambda(x_i) .
\]
%
Let $\beta>0$ be a parameter to be chosen later. By definition of
$\hat
\lambda$, we have for any $\lambda\in\Lambda_1^+$ that
%
%
\begin{equation}
\label{EQprMS1}
S(\hat\lambda) \ge S(\lambda) - \Delta_n(\lambda) -\beta\log M ,
\end{equation}
where $\Delta_n(\lambda)=2\sum_{i=1}^n(Y_i-f(x_i))\hh_{\hat\lambda
-\lambda}(x_i)-\beta\log M$.
The following lemma is useful to control the term $\Delta_n(\lambda)$
both in expectation and with high probability.
%
%
\begin{lem}
\label{LEMexpDelta}
Under Condition~\ref{cond1}, for any $\lambda\in\Lambda_1^+$ we have
\[
\E\Biggl[\exp\Biggl(\frac{ \Delta_n(\lambda)}{\beta}-\frac
{2B^2an}{\beta
^2}\sum_{j=1}^M\hat\lambda_j\|h_j-\hh_{ \lambda}\|^2\Biggr)
\Biggr]\le
1 .
\]
\end{lem}
\begin{pf}
For any $\lambda\in\Lambda_1^+$, $j=1, \ldots, M$,
define $\Upsilon_j$ by
\[
\Upsilon_j(\lambda)=\frac{2B^2an}{\beta^2}\|h_j-\hh_{ \lambda}\|
^2 .
\]
Jensen's inequality and the fact that $\log M=\sum_{j=1}^M \hat
\lambda
_j(\log M)$ yield
\begin{eqnarray*}
&&\E\Biggl[\exp\Biggl(\frac{ \Delta_n(\lambda)}{\beta}-\sum
_{j=1}^M\hat
\lambda_j\Upsilon_j(\lambda)\Biggr)\Biggr]\\
&&\qquad\le\E\Biggl[\sum_{j=1}^M \hat\lambda_j \exp\Biggl(\frac
{2}{\beta}\sum
_{i=1}^n\bigl(Y_i-f(x_i)\bigr)\bigl(h_j(x_i)-\hh_\lambda(x_i)\bigr)-\log
M-\Upsilon
_j(\lambda)\Biggr)\Biggr]\\
&&\qquad\le\frac{1}{M}\sum_{j=1}^M\E\Biggl[ \exp\Biggl(\frac{2}{\beta
}\sum
_{i=1}^n\bigl(Y_i-f(x_i)\bigr)\bigl(h_j(x_i)-\hh_\lambda(x_i)\bigr)-\Upsilon
_j(\lambda)\Biggr)\Biggr] .
\end{eqnarray*}
Now, from~(\ref{EQmgf}), which holds under Condition~\ref{cond1}, we
have for any $\lambda\in\Lambda_1^+$, $j=1, \ldots, M$, that
\[
\E\Biggl[\exp\Biggl(\frac{2}{\beta} \sum_{i=1}^n\bigl(Y_i-f(x_i)\bigr)
\bigl(h_j(x_i)-\hh_{\lambda}(x_i)\bigr)\Biggr)\Biggr]\le\exp
(\Upsilon
_j(\lambda) ) ,
\]
and the result of the lemma follows from the previous two displays.
\end{pf}

Take any $\bar\lambda\in\argmax_{\lambda\in\Lambda_1^+}S(\lambda
)$ and observe that Condition~\ref{cond2} together with a second-order
Taylor expansion of the function $S(\cdot)$ around $\bar\lambda$ gives
for any $\lambda\in\Lambda_1^+$
\[
S(\lambda)\le S(\bar\lambda)+[\nabla_{\lambda
}S(\bar\lambda)]^\top(\lambda-\bar\lambda)-\frac{n\kappa
^2}{2}\|
\hh_{\lambda}-\hh_{\bar\lambda}\|^2 ,
\]
where $\nabla_{\lambda} S({\bar\lambda})$ denotes the
gradient of $\lambda\mapsto S(\lambda)$ at $\bar\lambda$. Since
$\bar
\lambda$ is a maximizer of $\lambda\mapsto S(\lambda)$ over the set
$\Lambda_1^+$ to which $\lambda$ also belongs, we find\vadjust{\goodbreak} that $\nabla
_\lambda S({\bar\lambda})^\top(\lambda-\bar\lambda)\le0$ so that,
together with~(\ref{EQprMS1}), the previous display yields
%
%
\begin{equation}
\label{EQprMS2}
\frac{n\kappa^2}{2}\|\hh_{\hat\lambda}-\hh_{\bar\lambda}\|^2
\le
S({\bar\lambda})-S(\hat\lambda) \le\Delta_n(\bar\lambda) +\beta
\log
M .
\end{equation}
\begin{pf*}{Proof of Theorem~\ref{THMSagE}}
Using the convexity inequality $t\le e^t-1$ for any $t\in\R$,
Lemma~\ref{LEMexpDelta} yields
\[
\E[\Delta_n(\bar\lambda)]\le\beta\E\sum
_{j=1}^M\hat\lambda
_j\Upsilon_j(\bar\lambda)=\beta\E\sum_{j=1}^M\hat\lambda
_j\Upsilon
_j(\hat\lambda)+\frac{2B^2an}{\beta}\sum_{j=1}^M\E\|\hh_{\hat
\lambda
}-\hh_{\bar\lambda}\|^2 .
\]
The previous display combined with~(\ref{EQprMS2}) gives
\[
S({\bar\lambda})-\E[S(\hat\lambda)]\le\beta\E\sum
_{j=1}^M\hat\lambda_j\Upsilon_j(\hat\lambda)+\frac{4B^2a}{\beta
\kappa
^2}\bigl[S({\bar\lambda})-\E[S(\hat\lambda)]
\bigr]+ \beta
\log M
.
\]
It implies that for $\beta\ge8 B^2a/\kappa^2$
%
%
\begin{equation}
\label{EQprMS3}
S({\bar\lambda})-\E[S(\hat\lambda)] \le2\beta\E
\sum
_{j=1}^M\hat\lambda_j\Upsilon_j(\hat\lambda)+ 2\beta\log M
.
\end{equation}
Observe now that a second-order Taylor expansion of the function
$L(\cdot)$ around $\hat\lambda$, together with Condition~\ref{cond2},
gives for any $\lambda\in\Lambda_1^+$
\[
L(\lambda)\le L(\hat\lambda)+[\nabla_{\lambda
}L(\hat\lambda)]^\top(\lambda-\hat\lambda)-\frac{\kappa
^2}{2}\|
\hh_{\lambda}-\hh_{\hat\lambda}\|^2 .
\]
Thus
\[
\sum_{j=1}^M \hat\lambda_j L(e_j)\le L(\hat\lambda)-\frac{\kappa
^2}{2}\sum_{j=1}^M \hat\lambda_j \|h_{j}-\hh_{\hat\lambda}\|^2 .
\]
It follows that
\[
S(\hat\lambda)=n\sum_{j=1}^M\hat\lambda_j L(e_j) +nL(\hat\lambda
)\le
2nL(\hat\lambda)-\frac{n\kappa^2}{2}\sum_{j=1}^M \hat\lambda_j \|
h_{j}-\hh_{\hat\lambda}\|^2 .
\]
Combined with~(\ref{EQprMS3}), the above inequality yields
\[
S({\bar\lambda})-2n\E[ L(\hat\lambda)]\le\biggl(
2\beta
-\frac{\kappa^2\beta^2}{4B^2 a}\biggr)\E\sum_{j=1}^M\hat\lambda
_j\Upsilon_j(\hat\lambda) +2\beta\log M\le2\beta\log M
\]
for $\beta\ge8 B^2a/\kappa^2$.
Note that for any $j=1, \ldots, M$, $S({\bar\lambda}) \ge S(e_j)=2nL(e_j)$
so that from~(\ref{EQLKL}), we get
\[
a\E[\cE_{\mathrm{KL}}(\hh_{\hat\lambda}, \cV, \cH)]=\max
_{1\le
j\le M}L(e_j) - \E[ L(\hat\lambda)]\le\frac{\beta
}{n}\log
M .
\]
\upqed\end{pf*}\eject
\begin{pf*}{Proof of Theorem~\ref{THMSagP}}
From Lemma~\ref{LEMexpDelta} and a Chernoff bound, we get for any
$\lambda\in\Lambda_1^+$ and any $\delta>0$ that
\[
\p\Biggl[ \Delta_n(\lambda)-\frac{2B^2an}{\beta}\sum_{j=1}^M\hat
\lambda
_j\|h_j-\hh_{ \lambda}\|^2> \beta\log(1/\delta) \Biggr]\le\delta.
\]
Thus, the event $\cA_\lambda(\delta)=\{ \Delta_n(\lambda)\le\frac
{2B^2an}{\beta}\sum_{j=1}^M\hat\lambda_j\|h_j-\hh_{ \lambda}\|
^2+\beta
\log(1/\delta)
\}$
has probability greater than $1-\delta$. Theorem~\ref{THMSagP} follows
by applying the same steps as in the proof of Theorem~\ref{THMSagE} but
on the event $\cA_{\bar\lambda}(\delta)$ instead of in expectation.
\end{pf*}

\subsection{\texorpdfstring{Proofs of Theorems \protect\ref{THlinagE}--\protect\ref{THcvxagP}}{Proofs of Theorems 3.3--3.6}}


The following lemma exploits the strong convexity property stated in
Condition~\ref{cond2}.
%
%
\begin{lem}
\label{lemlemproof}
Let $\phi_1, \ldots, \phi_{D}$ be an orthonormal basis of the linear
span of the dictionary $\cH$. Let $\Lambda$ be a closed convex subset
of $\R^M$ or $\R^M$ itself and assume that $(\cH, \Lambda)$ satisfies
Condition~\ref{cond2}. Denote by $\lambda^*$ any maximizer of
the\vspace*{1pt}
function $\lambda\mapsto L(\lambda)$ over the set $\Lambda$. Then any
maximum likelihood estimator~${\hat\lambda_n}$ satisfies
%
%
\begin{equation}
\label{EQlemproof1}
\frac{\kappa^2}{2}\|\hh_{\hat\lambda_n}-\hh_{\lambda^*}\|^2\le
L({\lambda^*})-L({\hat\lambda_n})\le\frac{2}{\kappa^2}\sum
_{j=1}^{D}\zeta_j^2 ,
\end{equation}
where $\zeta_j=\frac{1}{n}\sum_{i=1}^nY_i\phi_j(x_i)-\langle f,
\phi
_j\rangle, j=1, \ldots, D$.
Moreover, if $\Lambda\subset\Lambda_1^+$ is a~closed convex set,
then ${\hat\lambda_n}$ satisfies
%
%
\begin{equation}
\label{EQlemproof2}
\frac{\kappa^2}{2}\|\hh_{\hat\lambda_n}-\hh_{\lambda^*}\|^2\le
L({\lambda^*})-L({\hat\lambda_n})\le\max_{1\le j\le M}|\xi_j|,
\end{equation}
where $\xi_j=\frac{1}{n}\sum_{i=1}^nY_ih_j(x_i)-\langle f,
h_j\rangle,
j=1, \ldots, M$.
\end{lem}
\begin{pf}
%
A second-order Taylor expansion of the function $L(\cdot)$ around~$\lambda^*$ gives for any $\lambda\in\Lambda$
\[
L(\lambda)\le L(\lambda^*)+[\nabla_{\lambda
}L(\lambda
^*)]^\top(\lambda-\lambda^*)-\frac{\kappa^2}{2}\|\hh
_{\lambda}-\hh
_{\lambda^*}\|^2 ,
\]
where we used Condition~\ref{cond2} and where $\nabla_{
\lambda} L({\lambda^*})$ denotes the gradient of $\lambda
\mapsto L(\lambda)$ at $\lambda^*$. Since $\lambda^*$ is a maximizer of
$\lambda\mapsto L(\lambda)$ over the set $\Lambda$ to which~$\lambda$
also belongs, we find that $\nabla_\lambda L({\lambda^*})^\top
(\lambda
-\lambda^*)\le0$ so that
%
%
\begin{equation}
\label{EQlbconcave}
L({\lambda^*})-L(\lambda)\ge\frac{\kappa^2}{2}\|\hh_{\lambda
}-\hh
_{\lambda^*}\|^2
\end{equation}
for any $\lambda\in\Lambda$, which gives the left inequalities
in~(\ref{EQlemproof1}) and~(\ref{EQlemproof2}).

Next, from the definition of $\hat\lambda_n$, we have
%
%
\begin{equation}
\label{EQproof1}
L({ \hat\lambda_n})\ge L({\lambda^*})+T_n(\lambda^*- \hat\lambda
_n ) ,\vadjust{\goodbreak}
\end{equation}
where
\[
T_n(\mu)=\frac{1}{n}\sum_{i=1}^nY_i\hh_\mu(x_i)-\langle f , \hh
_\mu
\rangle,\qquad \mu\in\R^M .
\]
Writing $\hh_\mu=\sum_{j=1}^{D}\nu_j \phi_j, \nu\in\R^D$, we
find that
\[
T_n(\mu)=\sum_{j=1}^{D}\nu_j\Biggl(\frac{1}{n}\sum_{i=1}^n Y_i \phi
_j(x_i)-\langle f, \phi_j\rangle\Biggr)=\sum_{j=1}^{D}\nu_j\zeta_j .
\]
Define the random variable
$
V_n= \sup_{\mu\in\R^M \dvtx\|\hh_\mu\|> 0}\{|T_n(\mu)
|/\|\hh
_\mu\|\} ,
$
so that $V_n$ satisfies
\[
V_n=\mathop{\sup_{\nu\in\R^M}}_{\nu\neq0}\frac{|{\sum
_{j=1}^{D}\nu_j\zeta_j}|}{(\sum_{j=1}^{D}\nu_j^2
)^{1/2}}=
\Biggl(\sum_{j=1}^{D}\zeta_j^2\Biggr)^{1/2} .
\]
Since $T_n(\lambda^*- \hat\lambda_n )\ge-V_n\|\hh_{\lambda^*-
\hat
\lambda_n }\|$, it yields together with~(\ref{EQproof1}) that
%
%
\begin{equation}
\label{EQubcs}
L( {\hat\lambda_n})\ge L({\lambda^*})-\|\hh_{\lambda^*- \hat
\lambda_n
}\|\Biggl(\sum_{j=1}^{D}\zeta_j^2\Biggr)^{1/2}.
\end{equation}
Combining~(\ref{EQubcs}) and~(\ref{EQlbconcave}) with $\lambda=\hat
\lambda_n$, we get~(\ref{EQlemproof1}).

We now turn to the proof of~(\ref{EQlemproof2}). From~(\ref{EQproof1}),
and the H\"older inequality, we have
%
\[
L({\lambda^*})-L({ \hat\lambda_n})\le\Biggl(\sum_{j=1}^M|\hat
\lambda
_{n,j} -\lambda^*_j|\Biggr) \max_{1\le j \le M}|\xi_j|\le\max
_{1\le j
\le M}|\xi_j| .
\]
Combined with~(\ref{EQlbconcave}), this inequality yields~(\ref{EQlemproof2}).
\end{pf}

In view of~(\ref{EQLKL}), to complete the proof of Theorems~\ref{THlinagE}--\ref{THcvxagP}, it is sufficient to bound from above the
quantities appearing on the right-hand side of~(\ref{EQlemproof1})
and~(\ref{EQlemproof2}). This is done using results from
Section~\ref{subexpfam} and by observing that the random variables $\zeta_j$ and
$\xi_j$ are of the form
%
%
\begin{equation}
\label{EQzetaj}
\zeta_j=S_n^{\omega^{(\zeta_j)}}-\E\bigl(S_n^{\omega^{(\zeta_j)}}\bigr)
,\qquad
\omega^{(\zeta_j)}_i=\frac{\phi_j(x_i)}{n} ,\qquad \bigl|\omega
^{(\zeta
_j)}\bigr|_2= \frac{1}{\sqrt{n}}\hspace*{-30pt}
\end{equation}
and
%
%
\begin{equation}
\label{EQxij}\qquad
\xi_j=S_n^{\omega^{(\xi_j)}}-\E\bigl(S_n^{\omega^{(\xi_j)}}\bigr)
,\qquad
\omega
^{(\xi_j)}_i=\frac{h_j(x_i)}{n} ,\qquad \bigl|\omega^{(\xi_j)}\bigr|_2\le
\frac
{R}{\sqrt n} ,\hspace*{-30pt}
\end{equation}
if $\max_{1\le j \le M}\|h_j\|\le R$.
%
\begin{pf*}{Proof of Theorem~\ref{THlinagE}}
Since the random variables $Y_i, i=1, \ldots, n$, are mutually
independent, we have
\[
\E[\zeta_j^2]= \operatorname{var}\Biggl(\frac{1}{n}\sum_{i=1}^n
Y_i\phi
_j(x_i)\Biggr)
\le\frac{\sigma^2}{n^2}\sum_{i=1}^n\phi_j^2(x_i)= \frac{\sigma
^2}{n} .
\]
Together with~(\ref{EQLKL}) and~(\ref{EQlemproof1}), this bound
completes the proof of Theorem~\ref{THlinagE}.
\end{pf*}
\begin{pf*}{Proof of Theorem~\ref{THlinagP}}
For any $s,t>0$, we have
\begin{eqnarray*}
\p\Biggl[\sum_{j=1}^D\zeta_j^2>t\Biggr]&=&
\p\Biggl[\frac{1}{D}\sum_{j=1}^D\zeta_j^2>\frac{t}{D}\Biggr]\le
e^{-{st}/{D}}\E\bigl[e^{ ({s}/{D})\sum_{j=1}^D\zeta
_j^2}\bigr]\\
&\le& e^{-{st}/{D}}\frac{1}{D}\sum_{j=1}^D\E[e^{ s\zeta
_j^2}]\le e^{-{st}/{D}}\frac{1}{D}\sum_{j=1}^D\sum
_{p=0}^\infty
\frac{s^p}{p!}\E[\zeta_j^{2p}] ,
\end{eqnarray*}
where we used, respectively: the Markov inequality, the Jensen
inequality and Fatou's lemma.
Observe now that~(\ref{EQmoment}), which holds under Condition~\ref{cond1}, and~(\ref{EQzetaj}) yield
\[
\E[\zeta_j^{2p}]\le C_{2p}\bigl|\omega^{(\zeta_j)}\bigr|_2^{2p}=\frac
{C_{2p}}{n^p}=2(p!)\biggl(\frac{2aB^2}{n}\biggr)^p .
\]
Therefore, the last two displays with $s=n/(4aB^2)$ yield
%
\[
\p\Biggl(\sum_{j=1}^D\zeta_j^2>t\Biggr)\le4e^{-{nt}/({4aB^2D})} .
\]
Theorem~\ref{THlinagP} follows by taking $t=\frac{4aB^2D}{n}\log
(4/\delta)$ in the previous display together with~(\ref{EQLKL})
and~(\ref{EQlemproof1}).
\end{pf*}

Before completing the proof of Theorems~\ref{THcvxagE} and~\ref{THcvxagP},
observe that~(\ref{EQmgf}) and~(\ref{EQxij}) imply that for any $j=1,
\ldots, M$, the random variable $|\xi_j|$ is sub-Gaussian with variance
proxy $\sigma^2=(RB)^2a/n$, that is,
%
%
\begin{equation}
\label{EQsubGaussxi}
\E\bigl[\mathrm{e}^{s|\xi_j|}\bigr] \le e^{{s^2\sigma
^2}/{2}}=e^{
{s^2(RB)^2a}/({2n})} .
\end{equation}
\begin{pf*}{Proof of Theorem~\ref{THcvxagE}}
%
It follows from Lemma 2.3 in \citet{Mas07} with the above choice
of variance proxy that
\[
\E\Bigl[{\max_{1\le j \le M}}|\xi_j| \Bigr]\le RB\sqrt{\frac{
a\log
M}{n}} .
\]
Combined with~(\ref{EQLKL}) and~(\ref{EQlemproof2}) the previous
inequality completes the proof of Theorem~\ref{THcvxagE}.
\end{pf*}
%
%
\begin{pf*}{Proof of Theorem~\ref{THcvxagP}}
Using, respectively, a union bound, a Chernoff bound and~(\ref{EQsubGaussxi}), we find
\[
\p\Bigl({\max_{1\le j \le M}}|\xi_j| >t \Bigr)\le M \exp\biggl( \frac
{nt^2}{2(RB)^2a}\biggr) .
\]
Together with~(\ref{EQLKL}) and~(\ref{EQlemproof2}), this bound
completes the proof of Theorem~\ref{THcvxagP} by taking $t=RB\sqrt
{\frac
{2a\log(M/\delta)}{n}}$.
\end{pf*}

\section*{Acknowledgments}

The author would like to thank Ramon
van Handel, Guillaume Lecu\'e and Vivian Viallon for helpful comments
and suggestions.

\begin{supplement}
\stitle{Minimax lower bounds}
\slink[doi]{10.1214/11-AOS961SUPP} 
\sdatatype{.pdf}
\sfilename{aos961\_supp.pdf}
\sdescription{Under some convexity and tail conditions, we prove
minimax lower bounds for the three problems of Kullback--Leibler
aggregation: model selection, linear and convex. The proof consists in
three steps: first, we identify a subset of admissible estimators, then
we reduce the problem to a usual problem of regression function
estimation under the mean squared error criterion and finally, we use
standard minimax lower bounds to complete the proof.}
\end{supplement}

%

\printaddresses

\end{document}